\documentclass[runningheads]{llncs}
\usepackage{graphicx}
\usepackage{amsmath}
\usepackage{amssymb}
\usepackage{multicol}
\usepackage{algorithm2e}
\usepackage{algpseudocode}
\usepackage{array}

\begin{document}
%
\title{Boundary-RL: Reinforcement Learning for Weakly-Supervised Prostate Segmentation in TRUS Images}
\titlerunning{Boundary-RL: Reinforcement Learning for Weakly-Supervised Segmentation}
%



\author{Weixi Yi\inst{1},
Vasilis Stavrinides\inst{2},
Zachary M.C. Baum\inst{1},
Qianye Yang\inst{1},
Dean C. Barratt\inst{1},
Matthew J. Clarkson\inst{1},
Yipeng Hu\inst{1},
Shaheer U. Saeed\inst{1}}
\authorrunning{W. Yi et al.}
%
\institute{Wellcome EPSRC Centre for Interventional and Surgical Sciences; Centre for Medical Image Computing; and Department of Medical Physics and Biomedical Engineering, University College London, London, UK \and
Division of Surgery \& Interventional Science, University College London; and Department of Urology, UCL Hospital NHS Foundation Trust, London, UK\\
\email{rmapwyi@ucl.ac.uk}}

\maketitle              
\begin{abstract}
We propose Boundary-RL, a novel weakly supervised segmentation method that utilises only patch-level labels for training. We envision the segmentation as a boundary detection problem, rather than a pixel-level classification as in previous works. This outlook on segmentation may allow for boundary delineation under challenging scenarios such as where noise artefacts may be present within the region-of-interest (ROI) boundaries, where traditional pixel-level classification-based weakly supervised methods may not be able to effectively segment the ROI. Particularly of interest, ultrasound images, where intensity values represent acoustic impedance differences between boundaries, may also benefit from the boundary delineation approach. Our method uses reinforcement learning to train a controller function to localise boundaries of ROIs using a reward derived from a pre-trained boundary-presence classifier. The classifier indicates when an object boundary is encountered within a patch, as the controller modifies the patch location in a sequential Markov decision process. The classifier itself is trained using only binary patch-level labels of object presence, which are the only labels used during training of the entire boundary delineation framework, and serves as a weak signal to inform the boundary delineation. The use of a controller function ensures that a sliding window over the entire image is not necessary. It also prevents possible false-positive or -negative cases by minimising number of patches passed to the boundary-presence classifier. We evaluate our proposed approach for a clinically relevant task of prostate gland segmentation on trans-rectal ultrasound images. We show improved performance compared to other tested weakly supervised methods, using the same labels e.g., multiple instance learning. 

\keywords{TRUS \and Weak Supervision \and Reinforcement Learning.}
\end{abstract}

\section{Introduction}

Automated segmentation of anatomical structures and other regions-of-interest (ROI) plays a vital role in the field of medical imaging \cite{sharma2010automated,ramesh2021review}. Accurate delineation and contouring of ROI boundaries is an essential step in procedures such as treatment planning e.g., for radiotherapy \cite{savjani2022automated}, longitudinal diagnostic or prognostic planning e.g., by tracking fetal dimensions \cite{zeng2021fetal} or organ sizes such as prostate volume \cite{xu2010information}, for locating abnormalities within imaged structures e.g., for lung abnormality detection \cite{dertkigil2022systematic}, and for trans-rectal ultrasound (TRUS) guided prostate biopsy or brachytherapy \cite{lei2019ultrasound}. In particular, of interest in this work, TRUS-guided biopsy has recently been aided by automated segmentation of the gland boundaries for precise sampling along the gland for accurate diagnoses \cite{lei2019ultrasound,wang2023prostate,liu20223}. Recent advancements in the field have been driven by the introduction of fully supervised learning-based segmentation methods \cite{wang2019deep,karimi2019accurate} that require expert annotated data for learning. However, due to the time-consuming nature of pixel-level annotation of medical images, the requirement for specialized knowledge and significant inter-observer variability, the acquisition of ample annotated data for model training may be challenging \cite{chen2020variability,czolbe2021segmentation,saeed2022image2,chalcroft2021development,saeed2021adaptable}. 

Weakly-supervised semantic segmentation (WSSS) may address some of the above-mentioned issues with fully-supervised learning by employing weak labels, such as image-level classification labels, for learning the segmentation task \cite{zhang2020survey,kervadec2019constrained,pathak2015constrained}. However, segmenting the boundaries of the prostate from TRUS images with weak labels remains significantly challenging \cite{hulsmans1995review} due to: (1) the often indistinct boundaries between the prostate and surrounding tissues caused by the low contrast of soft tissues, (2) the frequently missing segments of the prostate boundaries due to the presence of shadowing artefacts, (3) the considerable patient-dependent variability in prostate shape and size, (4) the uneven intensity distribution of the prostate, and (5) the lack of informative signals, e.g., in full supervision, where in commonly used approaches for weak supervision, presence of object to learn a valid prior such as global correlation between missing and distinct boundaries becomes challenging.

Within the literature on WSSS, methods based on multiple instance learning (MIL) have been proposed, with varying definitions of bags and patches used for training. For instance, MILinear \cite{ren2015weakly}, with its bag-splitting-based mechanism, is capable of iteratively generating new negative bags from positive ones. Xu et al. \cite{xu2014weakly} integrated the concept of clustering and introduced a multi-cluster instance learning technique for segmenting cancerous and non-cancerous tissues. CAMEL \cite{xu2019camel} treated lattice patches as instances and histopathological images as bags. Increasing research efforts are oriented towards considering the image as a bag and treating each pixel within the image as an instance, for improved efficiency at inference, thereby transforming the weakly supervised segmentation task with only image-level annotations into an instance prediction task based on bag-level annotations. Jia et al. \cite{jia2017constrained} proposed DWS-MIL, based on a fully convolutional network, for segmenting carcinogenic areas within histopathological images. Li et al. \cite{li2023weakly} introduced a self-attention mechanism to capture the correlation between instances within MIL. In addition to the MIL-based approaches, certain methods employ patch-level labels (i.e., binary labels of ROI presence within a patch or crop of a larger image) to achieve pixel-level segmentation with a comparatively lower annotation cost, as exemplified by the work of Han et al. \cite{han2022multi}. Most previously-proposed methods are application-specific with no general solutions being commonly utilised.

In this work, we present a novel WSSS approach that only utilises patch-level binary classification labels of object-presence for training. These are more-time efficient to obtain than full segmentation maps traditionally used in fully-supervised learning. The WSSS in this work is achieved by envisioning the segmentation task as a boundary detection problem rather than a pixel-level classification, which most previous works focus on. The boundary detection follows a Markov Decision Process (MDP) whereby a controller function, trained using reinforcement learning (RL) based on reward function derived from a boundary presence classifier (similar to previously proposed task-based rewards \cite{yoon2020data,saeed2022image,zhang2019adversarial,cubuk2018autoaugment}), aims to localise the boundary of the ROI.

The use of RL for boundary delineation allows us to pose this as a sequential MDP whereby the patch-location refinement is done until the boundary of the object is found, as determined by the boundary presence classifier. As opposed to previously proposed WSSS methods that pose the segmentation task as a pixel-level classification, our method does not require a sliding window to be passed over the entire image to generate pixel-level labels for the segmentation map, which may be more efficient at inference, reducing the number of forward passes through the classifier. Rather, the boundary delineation may simply be generated using a trajectory that the controller follows within an image. Moreover, this approach of boundary delineation has the potential to perform well even when artefacts such as shadowing are present within the ROI but not occluding the boundary, where traditional weakly supervised pixel-wise classification approaches to segmentation may produce inaccurate results \cite{pathak2014fully,pinheiro2015image,shi2020loss,kots2019u,ren2015weakly}.

Contributions of this work are summarised: First, we propose a RL-based framework, Boundary-RL, trained only using patch-level labels of object presence, capable of effectively identifying object boundaries; Second, we evaluate Boundary-RL for a clinically relevant task of prostate gland boundary delineation on TRUS images using data from real prostate cancer patients; And lastly, we compare Boundary-RL to commonly used algorithms such as fully-supervised learning and MIL-based WSSS, which we adapt for our application, from previous work.


\begin{figure}[]
    \centering
    \includegraphics[width=0.85\textwidth]{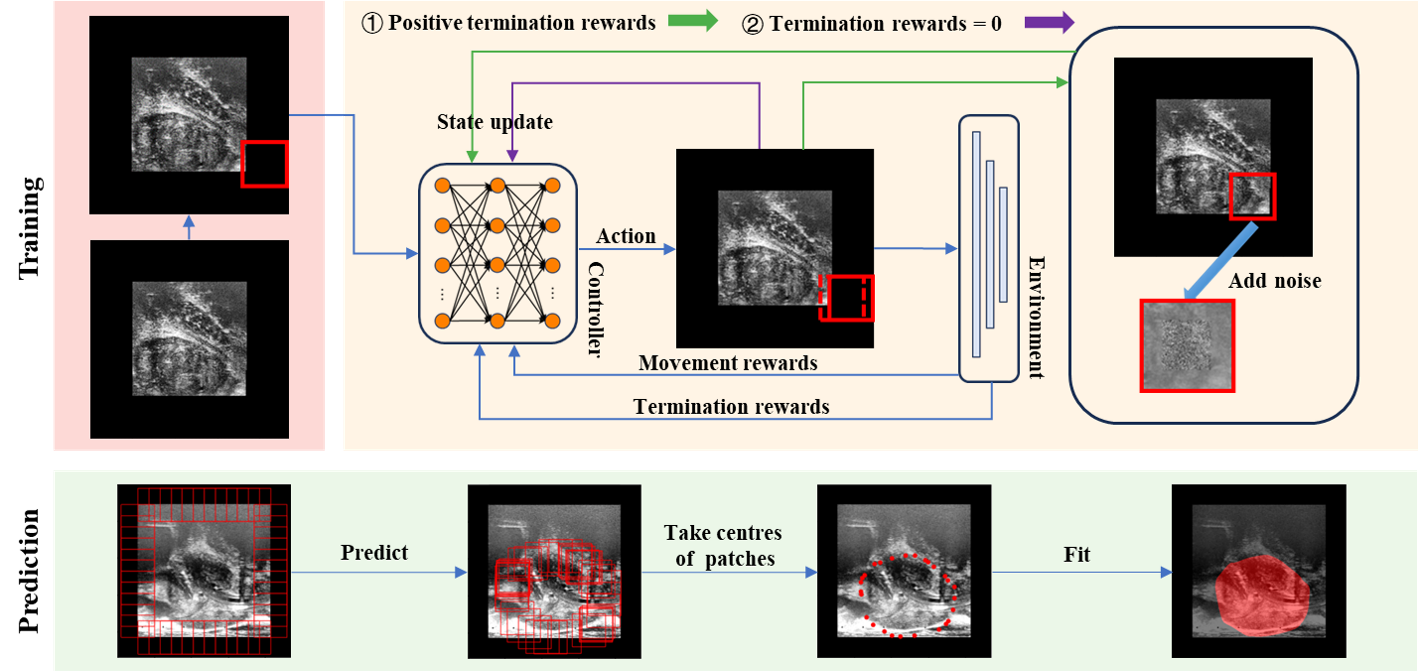}
    \caption{The workflow of our proposed framework.}
    \label{fig:workflow}
\end{figure}

\section{Methods}

In our work, the boundary presence classifier, trained using only binary patch-level labels of ROI presence within the patch, serves as a weak signal that may be used to train the boundary detecting controller. The patch-localising controller ensures that the location of the patch is refined, starting from the edge of the image, until part of the ROI exists within the patch, as determined by the boundary presence classifier. This may be repeated with randomised starting locations in order to get an effective boundary delineation for the ROI.

\subsection{Boundary presence classifier}

The boundary presence classifier $f(\cdot; w): \mathcal{X} \rightarrow \{0, 1\}$, with weights $w$, predicts object presence, given an image or image patch $x\in\mathcal{X}$, where $\mathcal{X}$ is the domain of images or image patches. Here, $0$, corresponds to object absence, and $1$, corresponds to object presence. Image-label pairs $\{x_k, y_k\}_{k=1}^N$ , where $y\in\{0, 1\}$, may be used to train the classifier $f$ using a binary cross-entropy loss function by means of gradient descent following $w^* = arg\min_w \mathbb{E} [L(y, f(x; w))]$ where $L(y, f(x; w))$ is the binary cross-entropy loss function
and $w^*$ are the optimal weights of the classifier which may be fixed after the initial pre-training stage. In practice the threshold for prediction probability is set as 0.9, so any classifier-predicted probability greater than 0.9 is rounded to 1, and 0 otherwise. The prediction probability of the classifier is treated as a boundary-presence probability prediction, which is used to inform the controller function, as outlined below.

\subsection{Boundary-RL - delineating boundaries of objects using RL}

We define an image patch $x_c$, of fixed dimensions, from image $x$ as having coordinates $c = (i, j)$. The aim is for the controller function is to refine $c$ until the ROI or its boundary are within the bounds of the patch. 

The controller function $g(\cdot; \theta): \mathcal{X} \times \mathcal{X} \rightarrow \mathcal{A}$, with weights $\theta$, takes in the image $x\in\mathcal{X}$, the image patch $x_c\in\mathcal{X}$ and outputs an action $a\in\mathcal{A}$ where $\mathcal{A}$ is the discrete action space defined as $a\in\mathcal{A}\in\{-1, 0, +1\}^2$. Here the two values for $a$ correspond to the change in coordinates for the patch, in each axis i.e., $a = (\delta i, \delta j)$ which may be applied to the coordinates $c_t=(i, j)$ at time-step $t$, to translate the patch to the new coordinates $c_{t+1} = (i+\delta i, j+\delta j)$. In practice these actions are modelled as four discrete actions of up, down, left or right, that is to say that if $\delta i \neq 0$ then $\delta j = 0$ and if $\delta j \neq 0$ then $\delta i = 0$.

\subsubsection{Interacting with an environment to learn boundary delineation}

The image $x_t$ and pre-trained presence classifier $f(\cdot; w^*)$ may be considered part of a MDP environment with which the controller interacts, i.e., as a MDP. The environment may be defined as $(\mathcal{S}, \mathcal{A}, p, r, \pi, \gamma)$. 

\textbf{States:} The observed state $s_t\in\mathcal{S}$ for this environment consists of the image $x_t$ and the image patch localised by the controller $x_{c, t}$, i.e., $s_t = \{x_t, x_{c, t}\}$, where $\mathcal{S}$ is the state space. 

\textbf{Actions:} The action $a_t\in\mathcal{A}$, where $\mathcal{A}$ is the action space, impacts the observed state at the subsequent time-step $t+1$ as $a_t = (\delta i, \delta j)$, which leads to $c_{t+1} = (i+\delta i, j+\delta j)$, which in-turn results in an updated patch, $x_{c, t+1}$, which forms part of the updated state.

\textbf{State transition distribution:} The state-action conditioned state transition distribution is given by $p:\mathcal{S} \times \mathcal{S} \times \mathcal{A} \rightarrow [0, 1]$. Assuming next state $s_{t+1}$ as input, given the current state $s_t$ and action $a_t$, the probability of the next state is denoted as $p(s_{t+1} | s_t, a_t)$.

\textbf{Rewards:} The reward function $r:\mathcal{S}\times\mathcal{A}\rightarrow \mathbb{R}$ takes the state-action pair as input and produces a reward $R_t = r(s_t, a_t)$, at time-step $t$. The reward consists of two components. 1) The first component is the movement reward $r_{mov}$, which is computed with the help of the estimated centroid of the prostate gland, given by coordinates $c_p = (i_p, j_p)$. The centroid of the prostate may be practically easy to obtain e.g., by means of pixel-intensity values or by manually labelling a subset of samples in roughly aligned images potentially utilising intensity-based registration. Then for time-step $t$, the L$^2$-norm between the location of the patch and the centroid of the prostate gland is given by $||c_t - c_p||$. We then define $r_{mov} = +1$ if $||c_t - c_p|| ~ - ~ ||c_{t-1} - c_p|| \leq 0$ and $r_{mov} = -1$ otherwise. 2) The second component is a termination reward $r_{term}$, given when the patch contains the ROI or its boundary, as determined by the boundary presence classifier function $f(\cdot; w^*)$, with pre-trained fixed weights $w^*$. We compute $r_{term} = f(x_{c, t}; w^*)$ for the termination reward, which is $f(x_{c, t}; w^*) = 1$ when the ROI is in the patch and $f(x_{c, t}; w^*) = 0$ otherwise. The two rewards may then be combined into our reward function $r(s_t, a_t) = r_{mov} + 100  r_{term}$, where $100$ is a scalar that allows weighting the two components. In this formulation, $r_{mov}$ may be considered as reward shaping and $r_{term}$ as the final sparse reward.

\textbf{Policy:} The probability of performing an action $a_t$ is given by the policy $\pi:\mathcal{S}\times\mathcal{A}\rightarrow [0, 1]$, e.g., for time-step $t$ the probability of performing action $a_t$ is $\pi(a_t|s_t)$. Sampling an action according to a policy is denoted as $a_t \sim \pi(\cdot)$. Following the state transition distribution for sampling next states, the policy for sampling actions and the reward function for generating corresponding rewards, we can collect states, actions and rewards over multiple time-steps $(s_1, a_1, R_1, \dots, s_T, a_T, R_T)$, also called trajectories. These are used to update our controller.

\textbf{Policy optimisation:} The policy is modelled as a parametric neural network $\pi(\cdot; \theta)$ with weights $\theta$. This neural network predicts parameters of a distribution from which to sample actions, which are the mean and standard deviation of a Gaussian distribution for continuous actions in this work \cite{schulman2017proximal}. The action is sampled from this predicted distribution $a\sim\pi(\cdot)$. We compute a cumulative reward $Q^{\pi(\cdot; \theta)}(s_t, a_t) = \sum^T_{k=0} \gamma^k R_{t+k}$ using a discount factor $\gamma$ for future rewards. We can then use gradient ascent to obtain optimal policy parameters: $\theta^* = arg\max_\theta \mathbb{E} [Q^{\pi(\cdot; \theta)}(s_t, a_t)]$.

\textbf{Episodic training and terminal signals:}
In practice, we collect trajectories by defining episode termination as finding a patch containing the organ boundary. More concretely, we terminate an episode when $f(x_{c, t}; w^*) = 1$, we may denote this time-step at termination as $t_{term}$. At $t_{term}$, we add random Gaussian noise over our patch within the image such that finding the same patch-location on the next episode within the same image does not lead to $f(x_{c, t}; w^*) = 1$. This encourages finding unique patches as boundary points for the ROI. We conduct $M$ episodes for each image until $M$ patches have been found as the object boundary. At each episode the patch location is randomly selected as an edge-pixel on one of the four edges, so the controller learns to move from the edge of the image to the boundary of the object within the image. In practice, we also terminate the episode after $T$ time-steps in case the boundary cannot be found after $T$ steps, where in our formulation, $T=1000$.

\textbf{Using controller-predicted patches for object boundary delineation:} The centre-points of all $M$ patches are taken to be the estimated boundary points for the object, outliers are removed via a simple distance-based outlier removal i.e., by rejecting points beyond a distance of 10 pixels from the mean of the boundary point locations. A polygon is constructed using the remainder of the point points which serves as the delineated boundary. With sufficiently densely sampled points, the nearest points are connected without and topological issues found in practice. The entire training algorithm is summarised in Algo. \ref{algo:rl_algo}, where at inference, the only difference is that rewards are not computed and the controller is not updated.

\begin{algorithm*}[!ht]
\SetAlgoLined
\KwData{Images to construct $s_t \in \mathcal{S}$}
\KwResult{Trained RL policy $\pi_{\theta^*}$.}
\BlankLine
\While{not converged}{
Randomly sample an image $x_t \in \mathcal{X}$\;
\For{$m\leftarrow 1$ \KwTo $M$}{
Start at $t = 0$\;
Randomly sample patch coordinates $c_0 = (i, j)$\;
Construct the state $s_t = \{x_t, x_{c, 0}\}$\;
Sample the action $a_0$ according to the policy $a_0 \sim \pi_\theta(a_0 | s_0)$\;
Compute object presence-based reward $R_0 = r(s_0, a_0)$\;
\BlankLine
\For{$t\leftarrow 1$ \KwTo $T$}{
Note: $t$ is now iterating starting at $t=1$\;
Given $a_{t-1} = (\delta i, \delta j )$, update patch coordinates $c_{t} = (i+\delta i, j+\delta j)$\;
Construct the state $s_t = \{x_t, x_{c, t}\}$\;
Sample the action $a_t$ according to the policy $a_t \sim \pi_\theta(a_t | s_t)$\;
Compute object presence-based reward $R_t = r(s_0, a_0)$\;
End if object presence detected i.e., $t = t_{term}$ if $f(x_{c, t}; w^*) = 1$; add Gaussian noise to patch to update $x_t$\;
}}
\BlankLine
Once $M$ trajectories of $R_{t=1:T}$ or $R_{t=1:t_{term}}$ collected, update RL function using gradient ascent}
\caption{Training procedure for Boundary RL.}
\label{algo:rl_algo}
\end{algorithm*}

\section{Experiments}

\subsection{Dataset}

During the (NCT02290561, NCT02341677) clinical trials, TRUS images from 249 patients were collected using a bi-plane transperineal ultrasound probe during manual/rotational positioning of a digital transperineal stepper, resulting in a total of 5185 prostate-containing images of size $403\times361$, which were center-cropped to $360\times360$. Ground truth labels were acquired through pixel-wise consensus among three researchers and verified by an expert radiologist. From 203 patients, we randomly selected 4000 images for controller training, and further cropped 80,000 $90\times90$ patches for MIL and Boundary Presence Classifier training. From the remaining 46 patients, we randomly chose 200 images each for validation and testing.



\subsection{Architecture and implementation details}

\textbf{Boundary presence classifier:} EfficientNet \cite{tan2019efficientnet} architecture was used,  with the Adam optimiser with a learning rate of 1e-4 and batch size 16.

\textbf{Boundary RL controller:} We trained the controller, which has 3 convolutional layers followed by 2 dense layers, for Boundary-RL using Proximal Policy Optimization (PPO) \cite{schulman2017proximal}, which took about 27 hours on a single Nvidia Tesla V100 GPU, with a learning rate set at 3e-4 and a batch size of 4096.

\subsection{Comparisons and baselines}

\textbf{MIL:} MIL is adapted from previous works \cite{pathak2014fully,pinheiro2015image,shi2020loss,kots2019u,ren2015weakly}, for our application of prostate gland segmentation. We treat each patch as a bag, and each pixel within a patch is considered an instance. We use a modified U-Net \cite{ronneberger2015u}, with global average pooling added before the full connection layer to aggregate instance probability distribution vectors into bag features. We adopt both partition (MIL-P) and sliding window (MIL-S) approaches for prediction. For overlapping predictions within the sliding window, we take their average.

\textbf{Fully supervised learning (FSL):} We chose the commonly used U-Net, trained using segmentation labels, to compare with our proposed WSSS method.

\section{Results}

\begin{figure}[]
    \centering
    \includegraphics[width=0.85\textwidth]{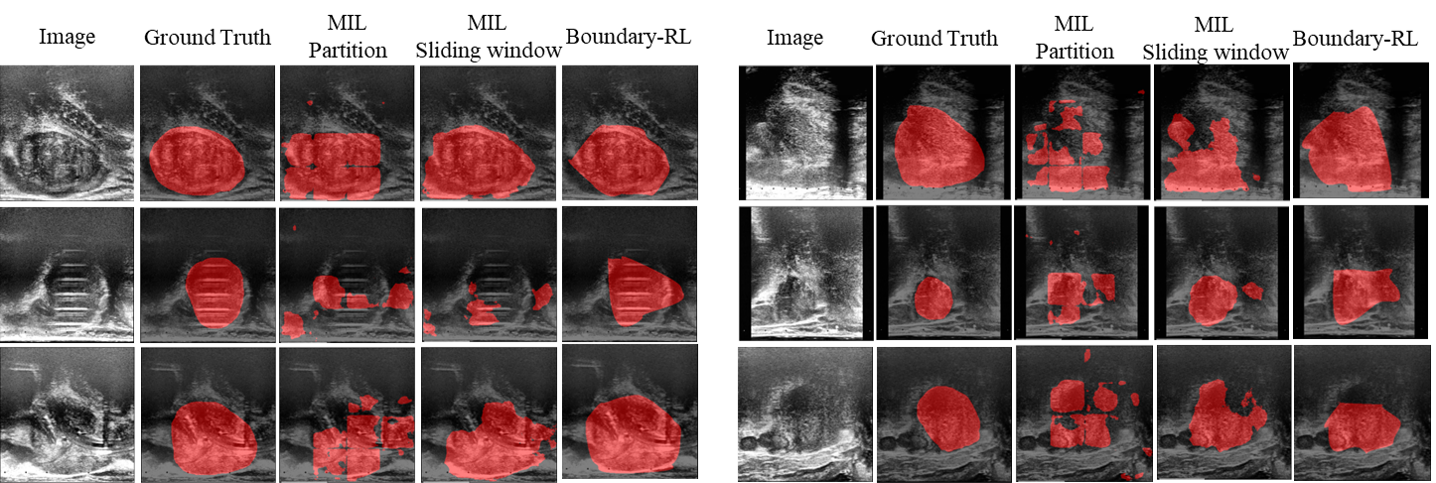}
    \caption{Samples with segmentation from the tested methods.}
    \label{fig:results}
\end{figure}

Results presented in Tab. \ref{tab:results}, show that our proposed Boundary-RL attained highest performance amongst the tested WSSS methods. While performance was statistically significantly better than the commonly used MIL with image partitioning (MIL-P) for inference, significance was not observed for the comparison with MIL using sliding windows (MIL-S) at inference. The fully supervised (FSL) method performed better than all tested WSSS methods, with statistical significance. 
Samples from each of the methods are presented in Fig. \ref{fig:results}.


\begin{table}[!ht]
\centering
\caption{Table of results and corresponding statistical tests.}
\begin{tabular}{|c|c| p{1.9cm}|p{1.9cm}|p{1.9cm}|p{1.9cm}|}
\hline
& & \multicolumn{4}{|c|}{Statistical tests (t-value/ p-value)}\\
\cline{3-6}
Algorithm & Dice & Proposed & MIL - P & MIL - S & FSL\\
\hline
Proposed & $0.751 \pm 0.161$ &             N/A         & -             &  -            & - \\
MIL - P & $0.663 \pm 0.156$ &         5.53/ $<$0.01  & N/A           & -             & - \\
MIL - S & $0.737 \pm 0.152$ &    0.88/ 0.38  & 4.80/ $<$0.01    & N/A           & - \\
FSL & $0.846 \pm 0.167$ &        5.76/ $<$0.01  & 11.26/ $<$0.01    & 6.79/ $<$0.01    & N/A \\
\hline
\end{tabular}
\label{tab:results}
\end{table}

\section{Discussion and Conclusion}

Results show that an effective boundary delineation policy may be learnt using our proposed Boundary-RL training scheme. Proposed adaptations to the commonly used MIL approach, also yielded comparable performance, with boundary RL being potentially more time-efficient at inference due to a fewer passes through the classifier due to the controller-learnt localisation policy, compared to sliding windows in MIL which may be akin to brute-search. All WSSS methods under-performed fully-supervised learning as expected due to the use of pixel-level segmentation maps for training, compared to only binary classification labels used in WSSS. Boundary RL outperformed all tested WSSS methods but statistical significance was not observed compared to MIL-S at inference. We maintain that our method produces an effective boundary delineation, evidenced from qualitative results and high observed Dice. Qualitative samples show that Boundary-RL can handle cases where artefacts exist within ROI boundaries, where MIL was unable to effectively segment since it relies on pixel-level classification which may be impacted by presence of artefacts within ROI boundaries. Future studies could explore augmenting rewards with other weak supervision signals such as self re-construction instead of just the patch-level classification.

We proposed a WSSS method based on RL that envisions segmentation as a boundary detection problem. A MDP is formulated to effectively learn to localise ROI boundaries using a reward based on boundary-presence within the localised region. The reward, generated using a classifier trained using binary patch-level labels of object presence, serves as a weak signal to inform boundary detection. We evaluate our approach for a clinically relevant task of boundary delineation of the prostate gland on TRUS, used during navigation in TRUS guided biopsy.

\section*{Acknowledgements}

This work was supported by the EPSRC grant [EP/T029404/1], Wellcome/EPSRC Centre for
Interventional and Surgical Sciences [203145Z/16/Z], and the International Alliance for Cancer Early Detection, an alliance between Cancer Research UK [C28070/A30912; 73666/A31378], Canary Center at Stanford University, the University of Cambridge, OHSU Knight Cancer Institute, University College London and the University of Manchester.

\newpage
\bibliographystyle{splncs04}
\bibliography{mybib}

%




\end{document}